\documentclass[10pt,twocolumn,letterpaper]{article}

\usepackage{acpr}
\usepackage{times}
\usepackage{epsfig}
\usepackage{graphicx}
\usepackage{subfigure}
\usepackage{amsmath}
\usepackage{amssymb}
\usepackage{floatrow}

\usepackage{todonotes}
\usepackage{cite}

\usepackage[pagebackref=true,breaklinks=true,letterpaper=true,colorlinks,bookmarks=false]{hyperref}

\acprfinalcopy % *** Uncomment this line for the final submission

\def\acprPaperID{89} % *** Enter the acpr Paper ID here

\ifacprfinal\fi
\begin{document}

%%%%%%%%% TITLE
\title{DASA: Domain Adaptation in Stacked Autoencoders using Systematic Dropout}

%\author{Abhijit Guha Roy\\
%%Department of Electrical Engineering\\
%Indian Institute of Technology Kharagpur, India\\
%{\tt\small abhi4ssj@gmail.com}
%\and
%Debdoot Sheet\\
%%Department of Electrical Engineering\\
%Indian Institute of Technology Kharagpur, India\\
%{\tt\small debdoot@ee.iitkgp.ernet.in}}

\author{Abhijit Guha Roy and Debdoot Sheet\\
Department of Electrical Engineering, Indian Institute of Technology Kharagpur, India\\
{\tt\small abhi4ssj@gmail.com, debdoot@ee.iitkgp.ernet.in }}
%\and
%Debdoot Sheet\\
%Indian Institute of Technology Kharagpur\\
%{\tt\small debdoot@ee.iitkgp.ernet.in}
%}

\maketitle
%\thispagestyle{empty}

%%%%%%%%% ABSTRACT
\begin{abstract}
Domain adaptation deals with adapting behaviour of machine learning based systems trained using samples in source domain to their deployment in target domain where the statistics of samples in both domains are dissimilar. The task of directly training or adapting a learner in the target domain is challenged by lack of abundant labeled samples. In this paper we propose a technique for domain adaptation in stacked autoencoder (SAE) based deep neural networks (DNN) performed in two stages: (i) unsupervised weight adaptation using systematic dropouts in mini-batch training, (ii) supervised fine-tuning with limited number of labeled samples in target domain. We experimentally evaluate performance in the problem of retinal vessel segmentation where the SAE-DNN is trained using large number of labeled samples in the source domain (DRIVE dataset) and adapted using less number of labeled samples in target domain (STARE dataset). The performance of SAE-DNN measured using $logloss$ in source domain is $0.19$, without and with adaptation are $0.40$ and $0.18$, and $0.39$ when trained exclusively with limited samples in target domain. The area under ROC curve is observed respectively as $0.90$, $0.86$, $0.92$ and $0.87$. The high efficiency of vessel segmentation with DASA strongly substantiates our claim.
\end{abstract}

%%%%%%%%% BODY TEXT
\begin{figure*}[t]
\begin{center}
%\missingfigure{Insert the graphical abstract here.}
  \includegraphics[width=0.95\linewidth]{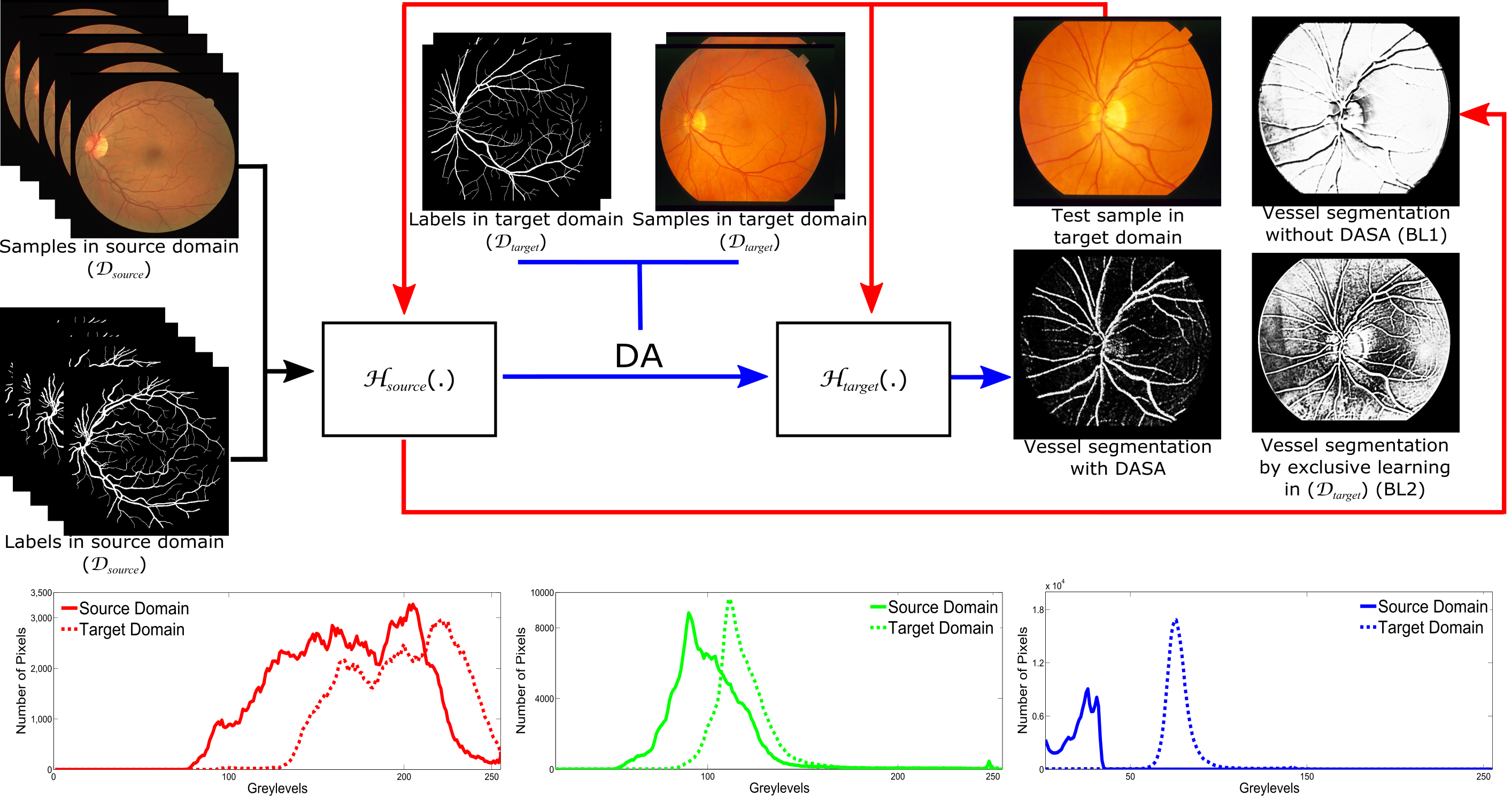}
\end{center}
   \caption{Overview of the process of DASA. It start with learning a SAE-DNN model $\mathcal{H}_\mathit{source}$ using ample labeled samples in $\mathcal{D}_\mathit{source}$. Limited number of labeled samples in $\mathcal{D}_\mathit{target}$ are used to transform $\mathcal{H}_\mathit{source} \overset{DA}{\rightarrow} \mathcal{H}_\mathit{target}$. Results of vessel segmentation with domain adaptation are compared with (BL1) SAE-DNN trained in $\mathcal{D}_\mathit{source}$ and deployed in $\mathcal{D}_\mathit{target}$ without DASA and (BL2) SAE-DNN trained in $\mathcal{D}_\mathit{target}$. The shifts in distribution of color statistics across samples in $\mathcal{D}_\mathit{source}$ and $\mathcal{D}_\mathit{target}$ are also illustrated.}
\label{fig:GA}
\end{figure*}

\section{Introduction}

The under-performance of learning based systems during deployment stage can be attributed to dissimilarity in distribution of samples between the \emph{source domain} on which the system is initially trained and the \emph{target domain} on which it is deployed. Transfer learning is an active field of research which deals with transfer of knowledge between the \emph{source} and \emph{target domains} for addressing this challenge and enhancing performance of learning based systems~\cite{pan2010survey}, when it is challenging to train a system exclusively in the \emph{target domain} due to unavailability of sufficient labeled samples. While domain adaptation (DA) have been primarily developed for simple reasoning and shallow network architectures, there exist few techniques for adapting deep networks with complex reasoning~\cite{deng2014autoencoder}. In this paper we propose a systematic dropout based technique for adapting a stacked autoencoder (SAE) based deep neural network (DNN)~\cite{bengio2009learning} for the purpose of vessel segmentation in retinal images~\cite{abramoff2010retinal}. Here the SAE-DNN is initially trained using ample number of samples in the \emph{source domain} (DRIVE dataset\footnote{http://www.isi.uu.nl/Research/Databases/DRIVE}) to evaluate efficacy of DA during deployment in the \emph{target domain} (STARE dataset\footnote{http://www.ces.clemson.edu/~ahoover/stare}) where an insufficient number of labeled samples are available for reliable training exclusively in the \emph{target domain}.

%-------------------------------------------------------------------------
\textbf{Related Work:} Autoencoder (AE) is a type of neural network which learns compressed representations inherent in the training samples without labels. Stacked AE (SAE) is created by hierarchically connecting hidden layers to learn hierarchical embedding in compressed representations. An SAE-DNN consists of encoding layers of an SAE followed by a target prediction layer for the purpose of regression or classification. With increase in demand for DA in SAE-DNNs different techniques have been proposed including marginalized training~\cite{chen2012marginalized}, via graph regularization~\cite{peng2013marginalized} and structured dropouts~\cite{yang2014fast}, across applications including recognizing speech emotion~\cite{deng2014autoencoder} to fonts~\cite{wang2015real}.

\textbf{Challenge:} The challenge of DA is to retain nodes common across \emph{source} and \emph{target domains}, while adapting the domain specific nodes using fewer number of labeled samples. Earlier methods~\cite{chen2012marginalized,peng2013marginalized,yang2014fast} are primarily challenged by their inability to re-tune nodes specific to the \emph{source domain} to nodes specific for \emph{target domain} for achieving desired performance, while they are able to only retain nodes or a thinned network which encode domain invariant hierarchical embeddings. 

\textbf{Approach:} Here we propose a method for DA in SAE (DASA) using systematic dropout. The two stage method adapts a SAE-DNN trained in the \emph{source domain} following (i) unsupervised weight adaptation using systematic dropouts in mini-batch training with abundant unlabeled samples in \emph{target domain}, and (ii) supervised fine-tuning with limited number of labeled samples in \emph{target domain}. The systematic dropout per mini-batch is introduced only in the representation encoding (hidden) layers and is guided by a saliency map defined by response of the neurons in the mini-batch under consideration. Error backpropagation and weight updates are however across all nodes and not only restricted to the post dropout activated nodes, contrary to classical randomized dropout approaches~\cite{srivastava2014dropout}. Thus having different dropout nodes across different mini-batches and weight updates across all nodes in the network, ascertains refinement of domain specific hierarchical embeddings while preserving domain invariant ones. 

The problem statement is formally introduced in Sec.~\ref{PS}. The methodology is explained in Sec.~\ref{Method}. The experiments are detailed in Sec.~\ref{expt}, results are presented and discussed in Sec.~\ref{res} with conclusion in Sec.~\ref{conc}.

%-------------------------------------------------------------------------
\section{Problem Statement}
\label{PS}

Let us consider a retinal image represented in the RGB color space as $\mathcal{I}$, such that the pixel location $\mathbf{x}\in\mathcal{I}$ has the color vector $\mathbf{c}(\mathbf{x})=\{r(\mathbf{x}), g(\mathbf{x}), b(\mathbf{x})\}$. $N(\mathbf{x})$ is a neighborhood of pixels centered at $\mathbf{x}$. The task of retinal vessel segmentation can be formally defined as assigning a class label $y\in\{\mathrm{vessel, background}\}$ using a hypothesis model $\mathcal{H}(\mathcal{I},\mathbf{x},N(\mathbf{x});\{\mathcal{I}\}_\mathrm{train})$. When the statistics of samples in $\mathcal{I}$ is significantly dissimilar from $\mathcal{I}_\mathrm{train}$, the performance of $\mathcal{H}(\cdot)$ is severely affected. Generally $\{\mathcal{I}\}_\mathrm{train}$ is referred to as the \emph{source domain} and $\mathcal{I}$ or the set of samples used during deployment belong to the \emph{target domain}. The hypothesis $\mathcal{H}(\cdot)$ which optimally defines \emph{source} and \emph{target domains} are also referred to as $\mathcal{H}_\mathit{source}$ and $\mathcal{H}_\mathit{target}$. DA is formally defined as a transformation $ \mathcal{H}_{\mathit{source}}\overset{DA}{\longrightarrow} \mathcal{H}_{\mathit{target}}$ as detailed in Fig.~\ref{fig:GA}.

\section{Exposition to the Solution}
\label{Method}

Let us consider the source domain as $\mathcal{D}_\mathit{source}$ with abundant labeled samples to train an SAE-DNN $(\mathcal{H}_\mathit{source})$ for the task of retinal vessel segmentation, and a target domain $\mathcal{D}_\mathit{target}$ with limited number of labeled samples and ample unlabeled samples, insufficient to learn $\mathcal{H}_\mathit{target}$ reliably as illustrated in Fig.~\ref{fig:GA}. $\mathcal{D}_\mathit{source}$ and $\mathcal{D}_\mathit{target}$ are closely related, but exhibiting distribution shifts between samples of the \emph{source} and \emph{target domains}, thus resulting in under-performance of $\mathcal{H}_\mathit{source}$ in $\mathcal{D}_\mathit{target}$ as also illustrated in Fig.~\ref{fig:GA}. The technique of generating $\mathcal{H}_\mathit{source}$ using $\mathcal{D}_\mathit{source}$, and subsequently adapting $\mathcal{H}_\mathit{source}$ to $\mathcal{H}_\mathit{target}$ via systematic dropout using $\mathcal{D}_\mathit{target}$ is explained in the following sections.

\subsection{SAE-DNN learning in the source domain}

AE is a single layer neural network that encodes the cardinal representations of a pattern $\mathbf{p}=\{p_k\}$ onto a transformed spaces $\mathbf{y}=\{y_j\}$ with $\mathbf{w}=\{w_{jk}\}$ denoting the connection weights between neurons, such that

\begin{equation}
	\mathbf{y} =f_\mathrm{NL}( [\mathbf{w}\textrm{~~}\mathbf{b}].[\mathbf{p}\textrm{~;~}1])
\label{eq:ae1}
\end{equation}

\noindent where the cardinality of $y$ denoted as $|\mathbf{y}| = J\times 1$, $|\mathbf{p}|=K \times 1$, $|\mathbf{w}|=J\times K$, and $\mathbf{b}$ is termed as the bias connection with $|\mathbf{b}|=J \times 1$. We choose $f_\mathrm{NL}(\cdot)$ to be a sigmoid function defined as $f_\mathrm{NL}(z)={1}/(1+\exp(-z))$. AE is characteristic with another associated function which is generally termed as the decoder unit such that

\begin{equation}
	\hat{\mathbf{p}} =f_\mathrm{NL}( [\mathbf{w}'\textrm{~~}\mathbf{b}'].[\mathbf{y}\textrm{~;~}1])
\label{eq:ea2}
\end{equation}

\noindent where $|\hat{\mathbf{p}}|=|\mathbf{p}|=K \times 1$, $|\mathbf{w}'|=K\times J$ and $|\mathbf{b}'|=K \times 1$. When $|\mathbf{y}|<<|\{\mathbf{p}_n\}|$, this network acts to store compressed representations of the pattern $\mathbf{p}$ encoded through the weights $\mathbf{W}=\{\mathbf{w},\mathbf{b},\mathbf{w}',\mathbf{b}'\}$. However the values of elements of these weight matrices are achieved through learning, and without the need of having class labels of the patterns $\mathbf{p}$, it follows unsupervised learning using some optimization algorithm~\cite{hay:2011}, viz. stochastic gradient descent.

\begin{equation}
	\{\mathbf{w},\mathbf{b},\mathbf{w}',\mathbf{b}'\} = \underset{\{\mathbf{w},\mathbf{b},\mathbf{w}',\mathbf{b}'\}}{\arg\min}\left(J(\mathbf{W})\right) 
\end{equation}

\noindent such that $J(\cdot)$ is the cost function used for optimization over all available patterns $\mathbf{p}_n\in\{\mathbf{p}(\mathbf{x}), \mathbf{x}\in\mathcal{I}\}$

\begin{equation}
	J(\mathbf{W}) = \sum_n\|\mathbf{p}_n-\hat{\mathbf{p}}_n\| + \beta|\rho-\hat\rho_n|
\label{eq:costfcn}
\end{equation}

\noindent where $\beta$ regularizes the sparsity penalty, $\rho$ is the imposed sparsity and $\hat{\rho}_n$ is the sparsity observed with the $n^\mathrm{th}$ pattern in the mini-batch.

The SAE-DNN consists of $L=2$ cascade connected AEs followed by a softmax regression layer known as the target layer with $\mathbf{t}$ as its output. The number of output nodes in this layer is equal to the number of class labels such that $|\mathbf{t}|=|\Omega|$ and the complete DNN is represented as

\begin{equation}
	\begin{array}{r c l}
	\mathbf{t} &=& f_\mathrm{NL}\left([\mathbf{w}_3\textrm{~~}\mathbf{b}_3].\left[f_\mathrm{NL}\left( [\mathbf{w}_2~\textrm{~~}\mathbf{b}_2].\left[f_\mathrm{NL}\left([\mathbf{w}_1\textrm{~~}\mathbf{b}_1].\right.\right.\right.\right.\right.\\
& & \left.\left.\left.\left.\left.\left[\mathbf{p}\textrm{~~}1\right]^T\right)\textrm{~~}1\right]^T\right)\textrm{~~}1\right]^T\right)
	\end{array}
	\label{eq:saednn}
\end{equation}

\noindent where $\{\mathbf{W}_1=\{\mathbf{w}_1,\mathbf{b}_1\},\mathbf{W}_2=\{\mathbf{w}_2,\mathbf{b}_2\}\}$ are the pre-trained weights of the network obtained from the earlier section. The weights $\mathbf{W}_3=\{\mathbf{w}_3,\mathbf{b}_3\}$ are randomly initialized and convergence of the DNN is achieved through supervised learning with the cost function

\begin{equation}
	J(\mathbf{W}) = \sum_m\|\mathbf{t}_m-\Omega_m\|
	\label{eq:supcost}
\end{equation}

\noindent during which all the weights $\mathbf{W}=\{\mathbf{W}_1=\{\mathbf{w}_1,\mathbf{b}_1\},\mathbf{W}_2=\{\mathbf{w}_2,\mathbf{b}_2\}, \mathbf{W}_3=\{\mathbf{w}_3,\mathbf{b}_3\}\}$ are updated to completely tune the DNN. 

\subsection{SAE-DNN adaptation in the target domain}

\begin{figure}[t]
\begin{center}
\includegraphics[width=0.6\textwidth]{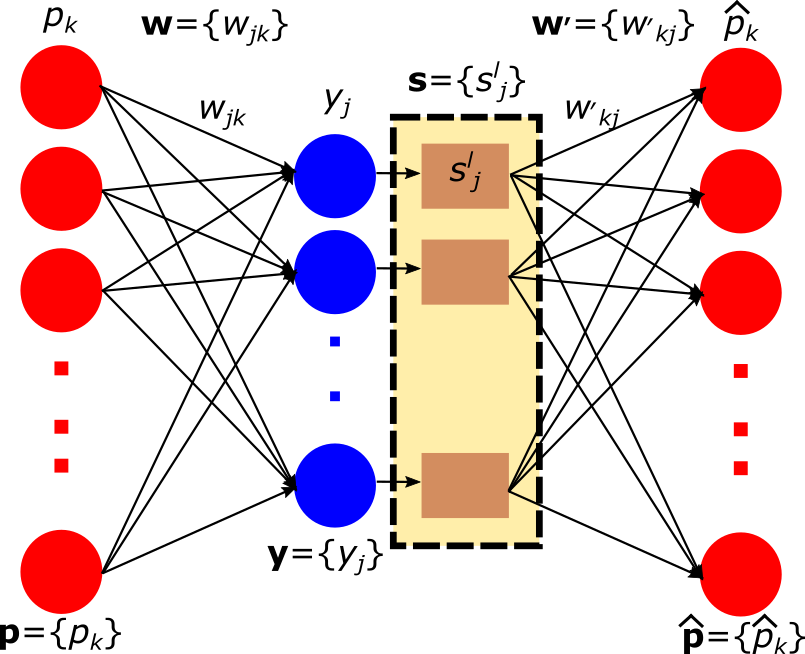}
\end{center}
   \caption{Illustration of the technique for unsupervised adaptation of SAE weights using systematic dropout.}
\label{fig:DropOut}
\end{figure}

\textbf{Unupervised adaptation of SAE weights using systematic dropouts}: The first stage of DA utilizes abundant unlabeled samples available in \emph{target domain} to retain nodes which encode domain invariant hierarchical embeddings while re-tuning the nodes specific in \emph{source domain} to those specific in \emph{target domain}. We follow the concept of systematic node drop-outs during training~\cite{srivastava2014dropout}. The number of layers and number of nodes in the SAE-DNN however remains unchanged during domain adaptation. Fig.~\ref{fig:DropOut} illustrates the concept.

Weights connecting each of the hidden layers is imported from the SAE-DNN trained in $\mathcal{D}_\mathit{source}$ are updated in this stage using an auto-encoding mechanism. When each mini-batch in $\mathcal{D}_\mathit{target}$ is fed to this AE with one of the hidden layers from the SAE-DNN; some of the nodes in the hidden layer exhibit high response with most of the samples in the mini-batch, while some of the nodes exhibit low response. The nodes which exhibit high-response in the mini-batch are representative of domain invariant embeddings which need to be preserved, while the ones which exhibit low-response are specific to $\mathcal{D}_\mathit{source}$ and need to be adapted to $\mathcal{D}_\mathit{target}$. We set a free parameter $\tau\in[0,1]$ defined as the transfer coefficient used for defining saliency metric $(\{s_j^l\}\in\mathbf{s})$ for the $j^{th}$ node in the $l^{th}$ layer as

\begin{equation}
	s_j^l=\begin{cases}
    1, & \text{if $y_j^l \geq \tau$}.\\
    0, & \text{otherwise}.
  \end{cases}
\end{equation}

\noindent where $y_j^l\in\mathbf{y}$ as in (\ref{eq:ae1}), and we redefine (\ref{eq:ea2}) while preserving (\ref{eq:costfcn}) and the original learning rules.

\begin{equation}
	\hat{\mathbf{p}} =f_\mathrm{NL}( [\mathbf{w}'\textrm{~~}\mathbf{b}'].[\mathbf{y}.\mathbf{s}\textrm{~;~}1])
\label{eq:ea22}
\end{equation}

\textbf{Supervised fine tuning with limited number of labeled samples:} The SAE-DNN with weight embeddings updated in the previous stage is now fine tuned using limited number of labeled samples in $\mathcal{D}_\mathit{target}$ following procedures in (\ref{eq:saednn}) and (\ref{eq:supcost}).

\section{Experiments}
\label{expt}

\textbf{SAE-DNN architecture:} We have a two-layered architecture with $L=2$ where AE$_1$ consists of $400$ nodes and AE$_2$ consists of $100$ nodes. The number of nodes at input is $15\times 15 \times 3$ corresponding to the input with patch size of $15 \times 15$ in the color retinal images in RGB space. AEs are unsupervised pre-trained with learning rate of $0.3$, over $50$ epochs, $\beta=0.1$ and $\rho=0.04$. Supervised weight refinement of the SAE-DNN is performed with a learning rate of $0.1$ over $200$ epochs. The training configuration of learning rate and epochs were same in the \emph{source} and \emph{target} domains, with $\tau=0.1$.

\textbf{Source and target domains:} The SAE-DNN is trained in $\mathcal{D}_\mathit{source}$ using $4\%$ of the available patches from the $20$ images in the training set in DRIVE dataset. DA is performed in $\mathcal{D}_\mathit{target}$ using (i) $4\%$ of the available patches in $10$ unlabeled images for unsupervised adaptation using systematic dropout and (ii) $4\%$ of the available patches in $3$ labeled images for fine tuning.

\textbf{Baselines and comparison:} We have experimented with the following SAE-DNN baseline (BL) configurations and training mechanisms for comparatively evaluating efficacy of DA: \textbf{BL1:} SAE-DNN trained in \emph{source domain} and deployed in \emph{target domain} without DA; \textbf{BL2:} SAE-DNN trained in \emph{target domain} with limited samples and deployed in \emph{target domain}.

\section{Results and Discussion}
\label{res}

\begin{figure*}[t]  
\begin{center}
	\subfigure[Source domain sample.]{\label{fig:dtls:src:photo} {\setlength\fboxsep{0pt}\setlength\fboxrule{0pt}\fbox{\includegraphics[height=0.14\textwidth]{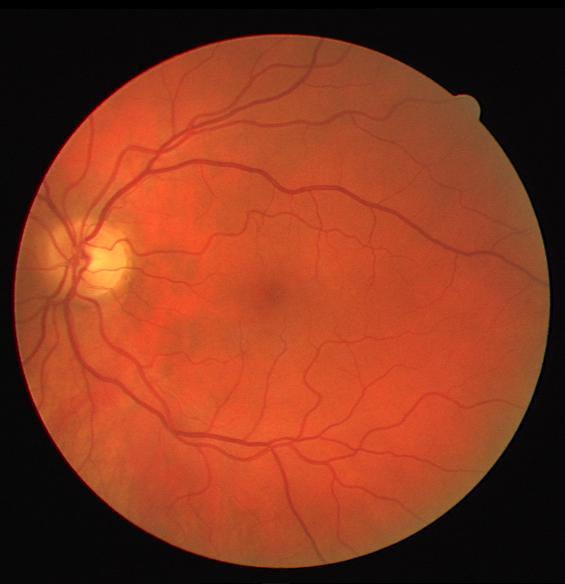}}}}
	\subfigure[Source domain labels.]{\label{fig:dtls:src:gt} {\setlength\fboxsep{0pt}\setlength\fboxrule{0pt}\fbox{\includegraphics[height=0.14\textwidth]{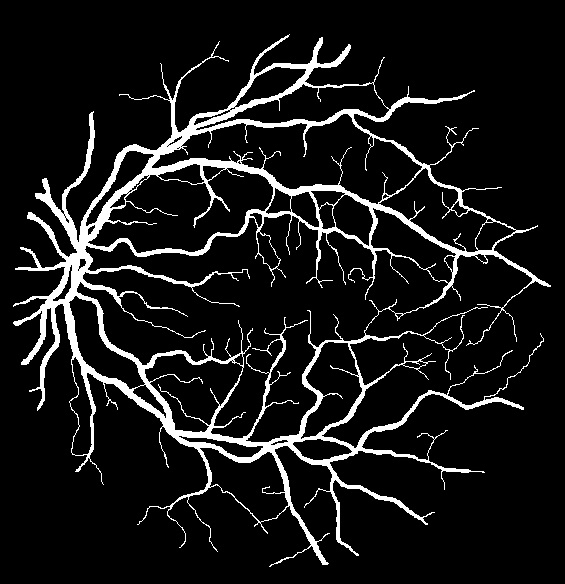}}}}
	\subfigure[Source domain prediction.]{\label{fig:dtls:src:pred} {\setlength\fboxsep{0pt}\setlength\fboxrule{0pt}\fbox{\includegraphics[height=0.14\textwidth]{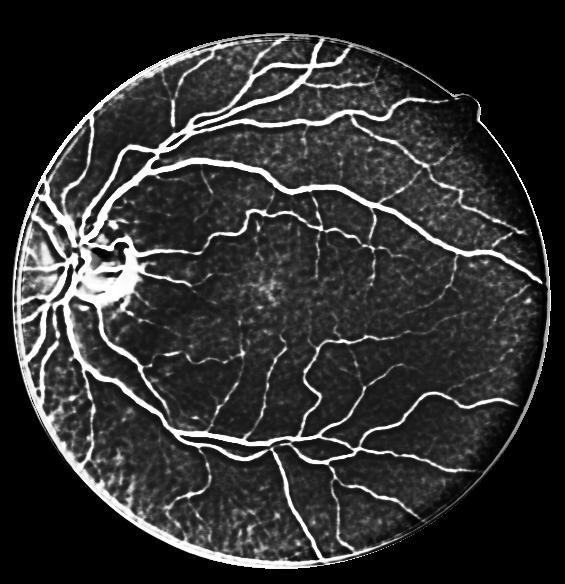}}}}
	\subfigure[Target domain sample.]{\label{fig:dtls:tgt:photo} {\setlength\fboxsep{0pt}\setlength\fboxrule{0pt}\fbox{\includegraphics[height=0.14\textwidth]{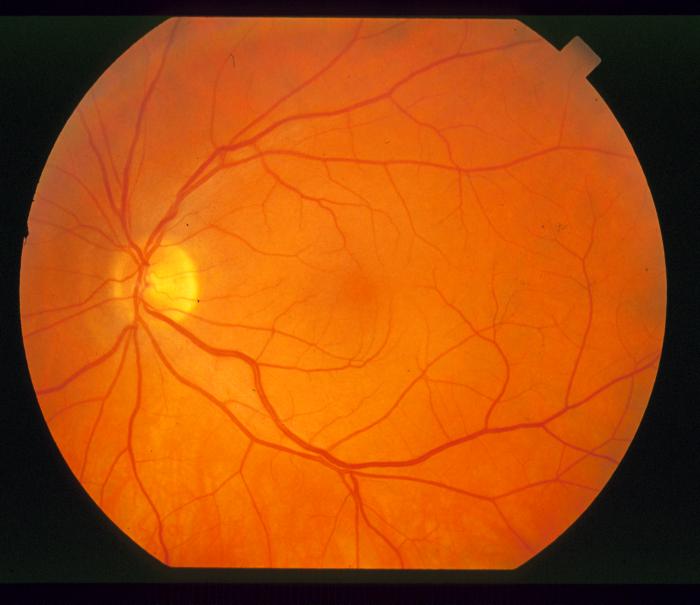}}}}
	\subfigure[Target domain labels.]{\label{fig:dtls:tgt:gt} {\setlength\fboxsep{0pt}\setlength\fboxrule{0pt}\fbox{\includegraphics[height=0.14\textwidth]{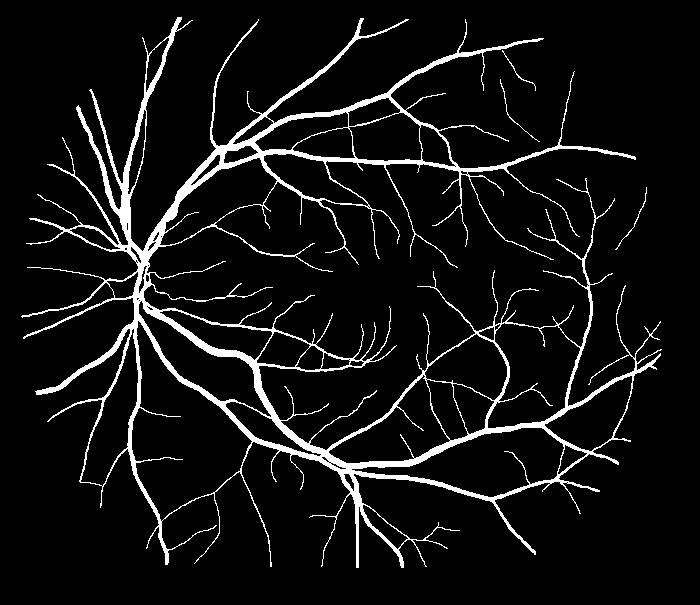}}}}
	\subfigure[Target domain prediction with DA.]{\label{fig:dtls:tgt:pred} {\setlength\fboxsep{0pt}\setlength\fboxrule{0pt}\fbox{\includegraphics[height=0.14\textwidth]{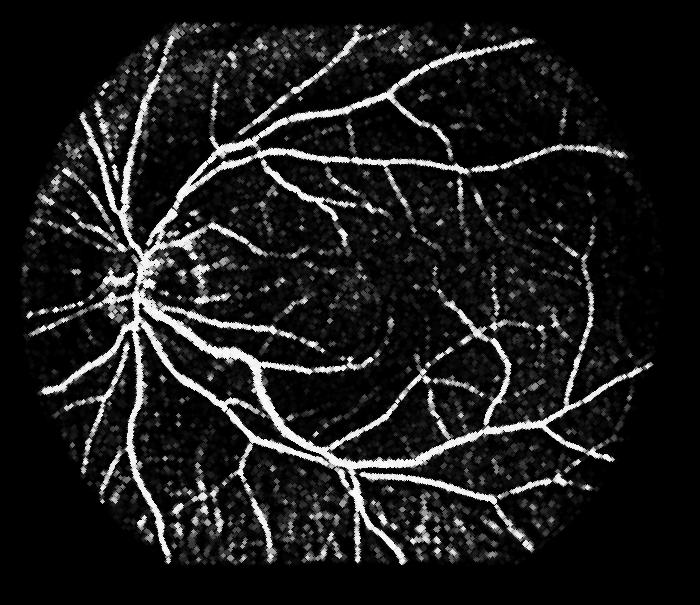}}}}
	\subfigure[Over complete representation in $\mathbf{w}_1$ in source domain.]{\label{fig:dtls:w1} {\setlength\fboxsep{0pt}\setlength\fboxrule{0pt}\fbox{\includegraphics[width=0.22\textwidth]{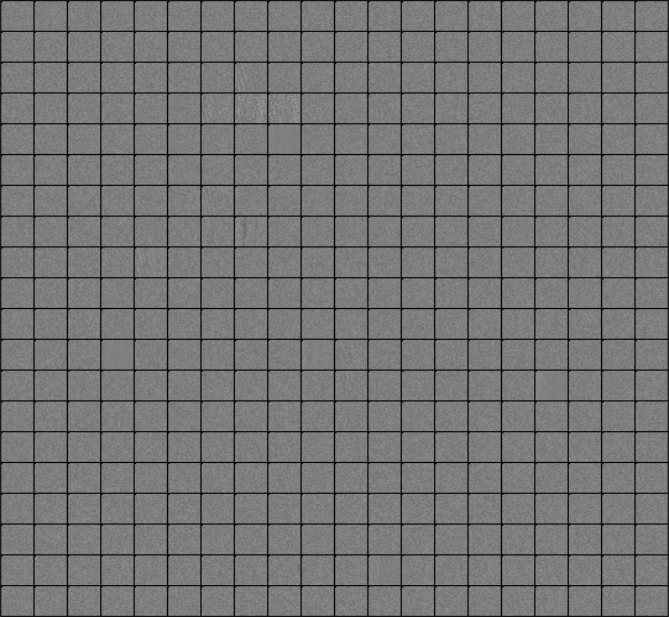}}}}
	\subfigure[Sparse representation in $\mathbf{w}_2$ in source domain]{\label{fig:dtls:w2} {\setlength\fboxsep{0pt}\setlength\fboxrule{0pt}\fbox{\includegraphics[width=0.22\textwidth]{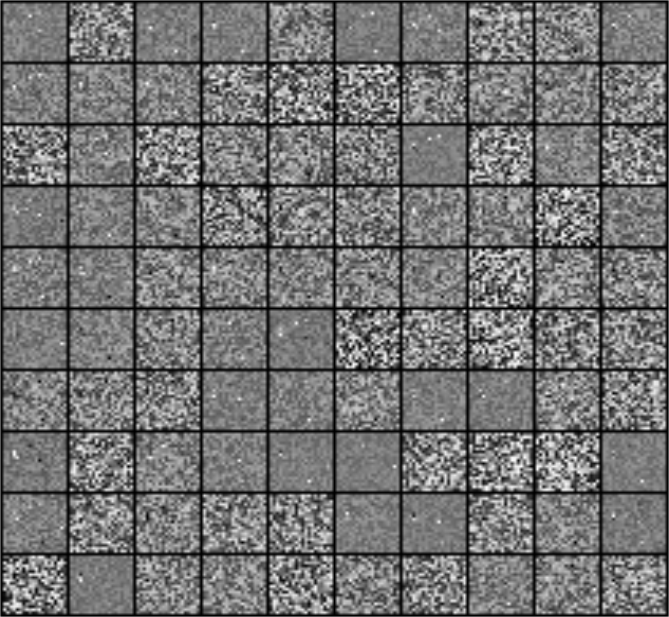}}}}
	\subfigure[DA representation in $\mathbf{w}_1$]{\label{fig:dtls:w1da} {\setlength\fboxsep{0pt}\setlength\fboxrule{0pt}\fbox{\includegraphics[width=0.22\textwidth]{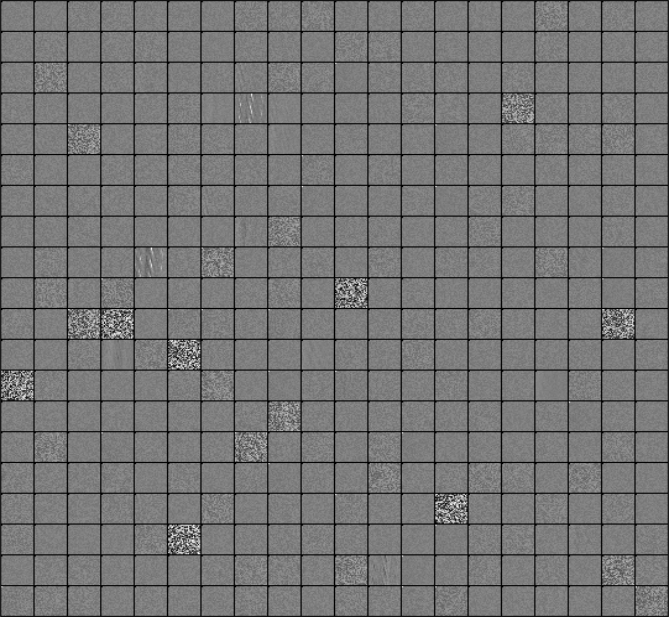}}}}
	\subfigure[DA representation in $\mathbf{w}_2$]{\label{fig:dtls:w2da} {\setlength\fboxsep{0pt}\setlength\fboxrule{0pt}\fbox{\includegraphics[width=0.22\textwidth]{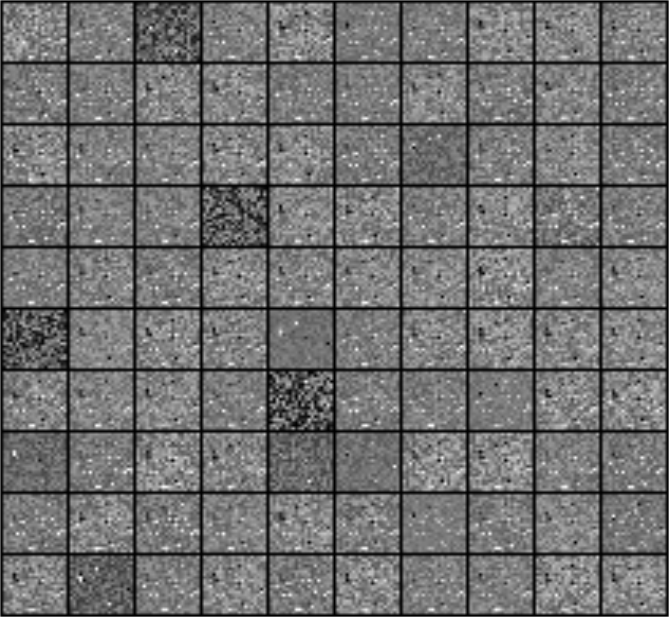}}}}
	\subfigure[$J(\mathbf{W})$ vs. epochs in training AE1]{\label{fig:dtls:llae1} {\setlength\fboxsep{0pt}\setlength\fboxrule{0pt}\fbox{\includegraphics[height=0.23\textwidth , width=0.3\textwidth]{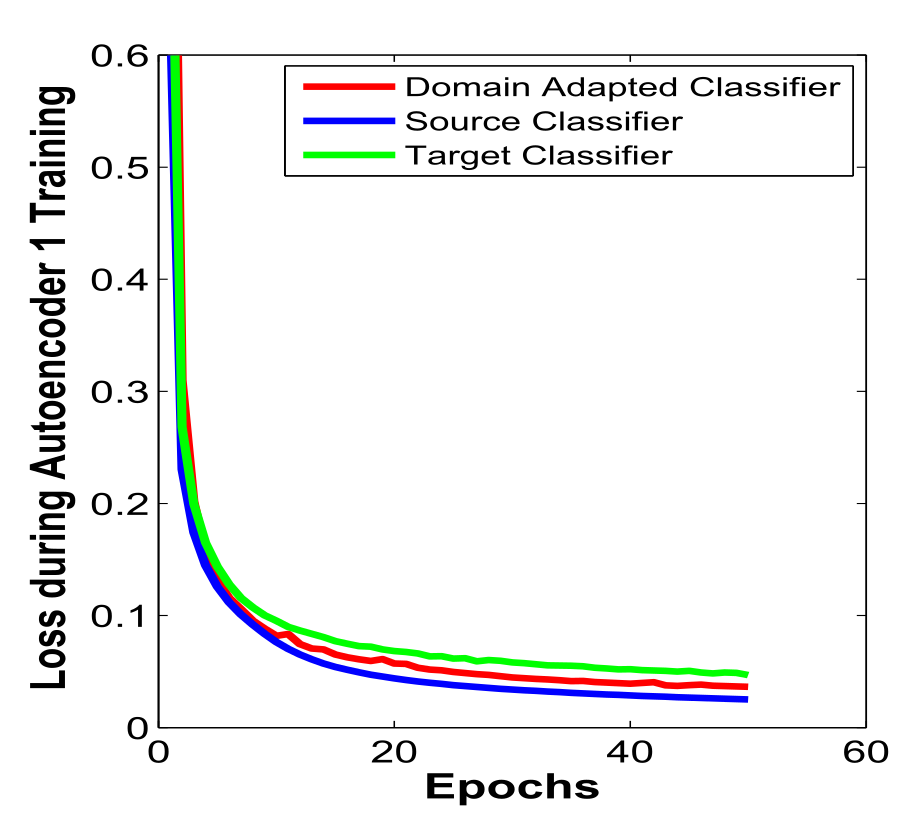}}}}
	\subfigure[$J(\mathbf{W})$ vs. epochs in training AE2]{\label{fig:dtls:llae2} {\setlength\fboxsep{0pt}\setlength\fboxrule{0pt}\fbox{\includegraphics[height=0.23\textwidth , width=0.3\textwidth]{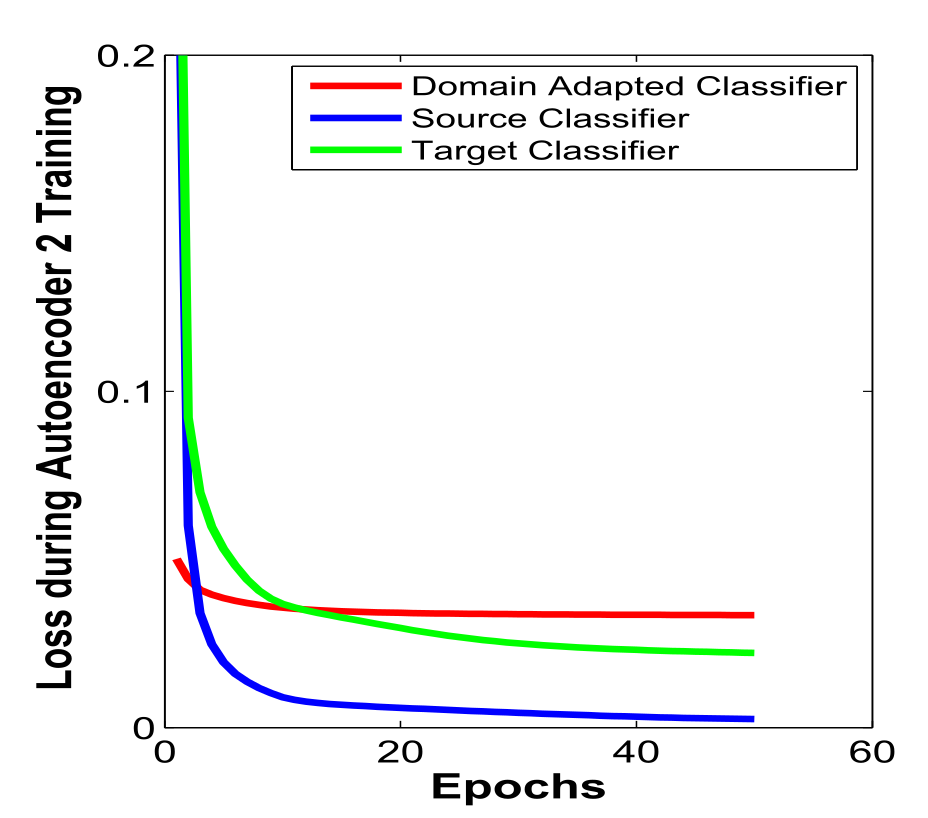}}}}
	\subfigure[$J(\mathbf{W})$ vs. epochs in training SAE-DNN]{\label{fig:dtls:dnn} {\setlength\fboxsep{0pt}\setlength\fboxrule{0pt}\fbox{\includegraphics[height=0.23\textwidth , width=0.3\textwidth]{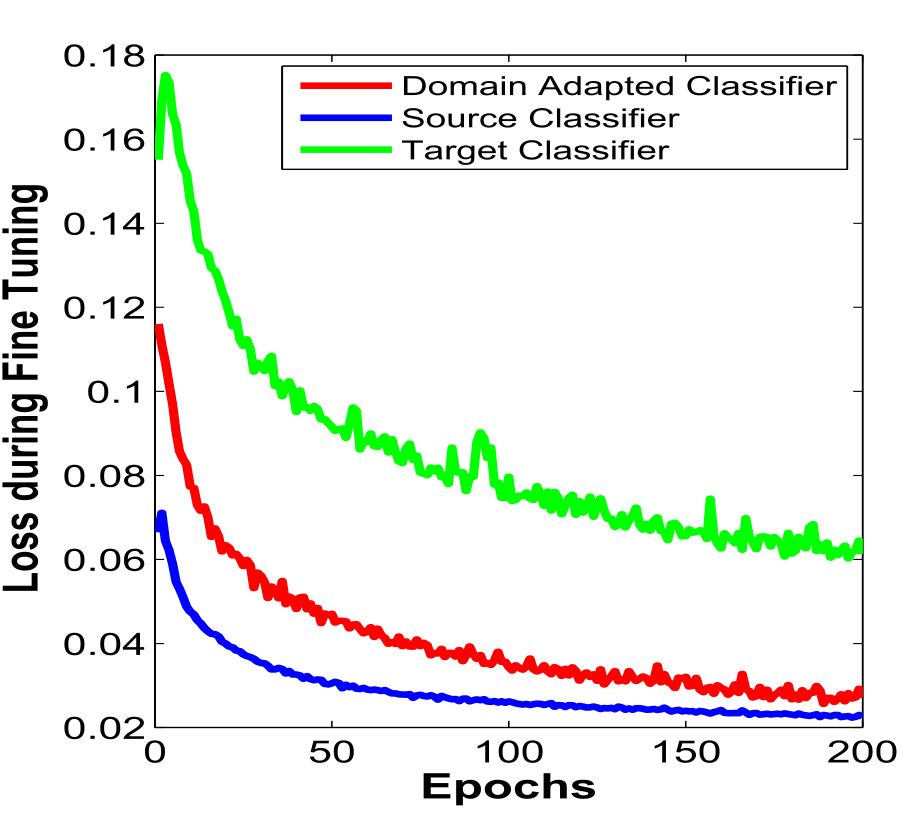}}}}
\end{center}
\caption{Performance of the vessel segmentation with (a-c) SAE-DNN on sample $01$ (Test) in $\mathcal{D}_\mathit{source}$ (DRIVE), (d-f) DASA on sample $0163$ in $\mathcal{D}_\mathit{source}$ (STARE), (g, h) representation learned by the SAE-DNN in $\mathcal{D}_\mathit{source}$, (i, j) DA representations using $\mathcal{D}_\mathit{target}$, learning dynamics vs. epochs in (k, l) AEs and (m) SAE-DNN indicating the higher efficacy of DASA compared to direct learning with limited samples in $\mathcal{D}_\mathit{target}$.}
\label{fig:dtls}
\end{figure*}

The results comparing performance of the SAE-DNN are reported in terms of $logloss$ and area under ROC curve as presented in Table~\ref{table:1}, and DA aspects in Fig.~\ref{fig:dtls}.

\begin{table}[!h]
\centering
\begin{tabular}{ |c|c|c| } 
 \hline
 & $logloss$ & Area under ROC \\ 
 \hline
 Source domain & $0.19\pm 0.05$ & $0.90\pm 0.02$ \\
 \hline
 BL1 & $0.40\pm 0.31$ & $0.86\pm 0.03$ \\ 
 \hline
 BL2 & $0.39\pm 0.68$ & $0.87\pm 0.01$ \\ 
 \hline
 \textbf{DASA} & $0.18\pm 0.02$ & $0.92\pm 0.02$ \\ 
 \hline
\end{tabular}
\caption{Comparison of Performance with the baselines}
\label{table:1}
\end{table}

\textbf{Hierarchical embedding in representations learned across domains:} AEs are typically characteristic of learning hierarchical embedded representations. The first level of embedding  represented in terms of $\mathbf{w}_1$ in Fig.~\ref{fig:dtls:w1} is over-complete in nature, exhibiting substantial similarity between multiple sets of weights which promotes sparsity in the nature of $\mathbf{w}_2$ in Fig.~\ref{fig:dtls:w2}. Some of these weight kernels are domain invariant, and as such remain preserved after DA as observed for $\mathbf{w}_1$ in Fig.~\ref{fig:dtls:w1da} and for $\mathbf{w}_2$ in Fig.~\ref{fig:dtls:w2da}. Some of the kernels which are domain specific, exhibit significant dissimilarity in $\mathbf{w}_1$ and $\mathbf{w}_2$ between \emph{source domain} in Figs.~\ref{fig:dtls:w1} and ~\ref{fig:dtls:w2} vs. \emph{target domain} in Figs.~\ref{fig:dtls:w1da} and ~\ref{fig:dtls:w2da}. These are on account of dissimilarity of sample statistics in the domains as illustrated earlier in Fig.~\ref{fig:GA} and substantiates DASA of being able to retain nodes common across \emph{source} and \emph{target domains}, while re-tuning domain specific nodes.

\textbf{Accelerated learning with DA:} The advantage with DA is the ability to transfer knowledge from \emph{source domain} to learn with fewer number of labeled samples and ample number of unlabeled samples available in the \emph{target domain} when directly learning in the \emph{target domain} does not yield desired performance. Figs.~\ref{fig:dtls:llae1} and \ref{fig:dtls:llae2} compare the learning of $\mathbf{w}_1$ and $\mathbf{w}_2$ using ample unlabeled data in \emph{source} and \emph{target domain} exclusively vs. DA. Fig.~\ref{fig:dtls:dnn} presents the acceleration of learning with DA in \emph{target domain} vs. learning exclusively with insufficient number of labeled samples.

\textbf{Importance of transfer coefficient:} The transfer coefficient $\tau$ drives quantum of knowledge transfer from the \emph{source} to \emph{target domains} by deciding on the amount of nodes to be dropped while adapting with ample unlabeled samples. This makes it a critical parameter to be set in DASA to avoid over-fitting and negative transfers as illustrated in Table.~\ref{tab:tau} where optimal $\tau=0.1$. Generally $\tau\in[0,1]$ with $\tau\rightarrow 0$ being associated with large margin transfer between domains when they are not very dissimilar, and $\tau\rightarrow 1$ being associated otherwise.

\begin{table}[!h]
\centering
\begin{tabular}{ |c|c|c|c|c|c| } 
 \hline
$\tau$ & $0$ & $0.05$ & $0.1$ & $0.15$ & $0.2$ \\ 
 \hline
 $logloss$ & $0.39$ & $0.24$ & $0.18$ & $0.21$ & $0.32$ \\
 \hline
\end{tabular}
\caption{Variation of $logloss$ in DA with variation of $\tau$}
\label{tab:tau}
\end{table}

\section{Conclusion}
\label{conc}

We have presented DASA, a method for knowledge transfer in an SAE-DNN trained with ample labeled samples in \emph{source domain} for application in \emph{target domain} with less number of labeled samples insufficient to directly train to solve the task in hand. DASA is based on systematic droupout for adaptation being able to utilize (i) ample unlabeled samples and (ii) limited amount of labeled samples in \emph{target domain}. We experimentally provide its efficacy to solve the problem of vessel segmentation when trained with DRIVE dataset (source domain) and adapted to deploy on STARE dataset (target domain). It is observed that DASA outperforms the different baselines and also exhibits accelerated learning due to knowledge transfer. While systematic drouput is demonstrated on an SAE-DNN in DASA, it can be extended to other deep architectures as well.

\section*{Acknowledgement}
We acknowledge NVIDIA for partially supporting this work through GPU Education Center at IIT Kharagpur.

\small
\bibliographystyle{ieee}
\bibliography{egbib}

\end{document}